\documentclass[10pt, a4paper]{article}
\usepackage{lrec}
\usepackage{multibib}
\newcites{languageresource}{Language Resources}
\usepackage{graphicx}
\usepackage{tabularx}
\usepackage{soul}

\usepackage{epstopdf}
\usepackage[latin1]{inputenc}
\usepackage{amsmath}
\usepackage{hyperref}
\usepackage{xstring}
\usepackage[figuresright]{rotating}

\usepackage{amssymb}

\usepackage{epstopdf}           
\usepackage{flushend}

\title{ \vspace*{.5\baselineskip} \textbf{Neural Machine Translation for Low-Resourced Indian Languages}}

\name{Himanshu Choudhary, Shivansh Rao, Rajesh Rohilla}

\address{Delhi Technological University (Formerly Delhi college of Engineering) \\
         himanshu.dce12@gmail.com, rao.shivansh570@gmail.com, rajesh@dce.ac.in\\}

\abstract{
A large number of significant assets are available online in English, which is frequently translated into native languages to ease the information sharing among local people who are not much familiar with English. However, manual translation is a very tedious, costly, and time-taking process. To this end, machine translation is an effective approach to convert text to a different language without any human involvement. Neural machine translation (NMT) is one of the most proficient translation techniques amongst all existing machine translation systems. In this paper, we have applied NMT on two of the most morphological rich Indian languages, i.e. English-Tamil and English-Malayalam. We proposed a novel NMT model using Multihead self-attention along with pre-trained Byte-Pair-Encoded (BPE) and MultiBPE embeddings to develop an efficient translation system that overcomes the OOV (Out Of Vocabulary) problem for low resourced morphological rich Indian languages which do not have much translation available online. We also collected corpus from different sources, addressed the issues with these publicly available data and refined them for further uses. We used the BLEU score for evaluating our system performance. Experimental results and survey confirmed that our proposed translator (24.34 and 9.78 BLEU score) outperforms Google translator (9.40 and 5.94 BLEU score) respectively. \\ \newline  }

\begin{document}

\maketitleabstract

\section{Introduction}

Many populated countries such as India and China have several languages which change region by region. for example, India has 23 constitutionally recognized official languages (\emph{e.g.}, Hindi, Malayalam, Telugu, Tamil, and Punjabi) and numerous unofficial local languages. Not only big countries, even small countries also rich in language diversity.  There are 851 languages spoken in Papua New Guinea, which is one of the smallest populated regions. In India, the population is about three billion, but only about 10\% of them can speak English\footnote{https://www.bbc.com/news/magazine-20500312}. Some studies say that out of those 10\% English speakers only 2\% can talk, write, and examine English well, and rest 8\% can merely recognize simple English and talk with a variety of accents. Thinking about a large number of valuable sources is available on the web in English and most people in India can not understand it well, it becomes important to translate such content into neighborhood languages to facilitate people. Sharing pieces of information between human beings is important not only for business purposes but also for sharing their emotions, reviews, and acts. For this, translation plays an essential role in minimizing the communication hole between different peoples. considering the vast amount of text, it is not viable to translate them manually. Hence, it becomes crucial to translate text from one language (say, English) to other languages (say, Tamil, Malayalam) automatically. This technique is also referred to as \emph{machine translation}.\\

English to Indian language translation poses the challenge of morphological and structural divergence. For instance, (i) the number of parallel corpora and  (ii) differences between languages, mainly the morphological richness and variation in word order due to syntactical divergence. Indian languages (IL) suffers from both of these problems, especially when they are being translated from English. Moreover, Indian languages such as Malayalam and Tamil differ not only in word order but are also more agglutinative as compared to English which is fusional. For instance, English has Subject-Verb-Object (SVO) whereas Tamil and Malayalam have Subject-Object-Verb (SOV). While syntactic differences contribute to difficulties of translation models,  morphological differences contribute to data sparsity. We attempt to overcome both issues in this paper.\\

There are various papers on machine translation, but apart from foreign languages most of the works on Indian languages are limited to Hindi and on conventional machine translation techniques such as \cite{smt} and \cite{translation}. Most of the previous work is focused on separating the words in suffix and prefix based on some rules and then applying translation techniques. We addressed this issue with BPE to make this whole process more efficient and reliable. Moreover, We observed that very less work is being done on low resourced Indian languages and techniques such as Byte-pair-encoding (BPEmb), MultiBPEmb, word-embedding, and self-attention are still unexplored which have shown a significant improvement in Natural Language Processing. Though unsupervised machine translation \cite{DBLP} is also in the focus of many researchers, still it is not as precise as supervised learning. We, also addressed that there is no trustworthy Public data available for the translation of such languages. Thus, in this paper, we have applied a neural machine translation technique with Multihead-self attention along with word embeddings and Pre-Trained Byte-Pair-Encoding. We worked on English-Tamil and English-Malayalam language pairs as it is one of the most difficult languages pair \cite{zdenekvzabokrtsky2012morphological} to translate due to morphological richness of Tamil and Malayalam language. A similar approach can be applied to other languages as well. We obtained the data from EnTamv2.0, Opus and UMC005, preprocessed them and evaluated our result using the evaluation matric BLEU. We used OpenNMT-py for the implementation of our models \footnote{http://opennmt.net/OpenNMT-py/}. Experimental results, as well as the survey by native peoples, confirms that our result is far better than conventional translation techniques on Indian languages.

The Main contributions of our work are as follows:
\begin{itemize}
\item This is the first work to apply pre-trained BPE and MultiBPE embeddings on Indian language pairs (English-Tamil, English-Malayalam) along with Multihead self-attention technique.
\item We achieved good accuracy with a relatively simpler model and in less training time rather than training on a complex neural network which requires much resources and time to train.
\item We have addressed the issues with data preprocessing of Indian languages and shown why it is a crucial step in neural machine translation.
\item We made our preprocessed data publicaly available,  which by our knowledge contains the largest number of a parallel corpus for the languages (English-Tamil, English-Malayalam, English-Telugu, English-Bengali, English-Urdu)  
\item Our model outperforms Google translator with a margin of 3.36 and an 18.07 BLEU score.
\end{itemize}

The paper is organized as follows. Sections Background and Approach describe the related work and the method that we used for our translator, respectively. Section experiments and Results show data preprocessing and results and analysis of our model. Finally, Section \ref{conclusion} concludes the paper and future work.

\section{Background}

A large amount of work has been reported on machine translation (MT) in the last few decades, the first one in the 1950s \cite{booth1955machine}. Various approaches is used by researchers, such as rule-based \cite{ghosh2014translation}, corpus-based \cite{wong2006machine}, and hybrid-based approach \cite{salunkhe2016hybrid}. Each approach has its own flaws and strength. Rule-based machine translation (RBMT) is MT systems based on the linguistic information about the source and target languages which is retrieved from ( multilingual, bilingual or monolingual) dictionaries and grammars covering the main syntactic, semantic and  morphological regularities. It is further divided into transfer-based approach (TBA)\cite{shilon2011} and inter-lingual based approach (IBA). In the Corpus-based approach, we use a large-sized parallel corpus as raw data. This raw data contains ground truth translation for the desired languages. These corpora are used to train the model for translation. A corpus-based approach further classified in (i) statistical machine translation (SMT) \cite{smt} and (ii) example-based machine translation (EBMT) \cite{somers2003overview}. SMT is the combination of decoding algorithms and basic statistical language models.EBMT, on the other hand, uses the translation examples and generates the new translation accordingly. It is done by finding the examples which are matching with the input. The alignment has to be performed after that to find out the parts of translation that can be reused.
Hybrid-base machine translation combines any corpus-based approach and transfer approach in order to overcome their limitations. According to the recent research \cite{khan2017} the machine translation performance of Indian languages such as  (\emph{e.g.}, Hindi, Bengali, Tamil, Punjabi, Gujarati, and Urdu) is of an average of 10\% accuracy. This demands the necessity of building better translation systems for Indian languages.
\newline
Unsupervised machine translation is further a new way of translation without using the parallel corpus, but the results are still not remarkable. On the other hand, NMT is an emerging technique and shown significant improvement in the translation results. In this paper \cite{hans2016improving} phrase-based hierarchical model is used and trained after morphological preprocessing. \cite{patel2017mtil17} trained their model after compound splitting and suffix separation. Many researchers also tried the same way and achieved a decent result on their respective datasets \cite{pathakneural}. We observed that morphological pre-processing, compound splitting and suffix or prefix separation can be overcome by using Byte-Pair-Encoding and produce similar or even better translation results without making the model complex.

\section{Approach}
In this paper, we present a neural machine translation technique using Multihead self-attention and word-embedding along with pre-trained Byte-Pair-Encoding (BPE) on our preprocessed dataset of Indian languages. We developed an efficient translation system, that overcomes the OOV (Out Of Vocabulary) and morphological analysis problem for Indian languages which do not have many translations available on the web. first, we provide an overview of NMT, Multi-head self-attention, word embedding, and Byte Pair Encoding. Next, we describe the framework of our translation model.

\subsection{Neural Machine Translation Overview}

Neural Machine translation is a powerful algorithm based on neural networks and uses the conditional probability of translated sentences to predict the target sentences of given source language \cite{revanuru2017neural}. When coupled with the power of attention mechanisms, this architecture can achieve impressive results with different variations. The following sub-sections provide an overview of basic sequence to sequence architecture, self-attention and other techniques that are used in our proposed translator. 

\subsubsection{Sequence to sequence architecture}
Sequence to sequence architecture is used for response generation whereas in Machine Translation systems it is used to find the relations between two language pairs. It consists of two important parts, an encoder, and a decoder. The encoder takes the input from the source language and the decoder leads to the output based on hidden layers and previously generated vectors.
Let $A$ be the source and $B$ be a target sentence. The encoding part converts the source sentence  $a_{1},a_{2},a_{3}...,a_{n}$ into the vector of fixed dimensions and the decoder part gives the word by word output using conditional probability. 
Here, $A_{1},A_{2},...,A_{M}$ in the equation are the fixed size encoding vectors. Using chain rule, the Eq.~\ref{eq1} is transformed to the Eq.~\ref{eq2}.

\begin{equation}
\begin{aligned}
P(B/A) = P(B|A_{1}, A_{2}, A_{3},...,A_{M})
\end{aligned}
\label{eq1}
\end{equation}

\begin{equation}
\begin{aligned}
P(B|A) =
P(b_{i}|b_{0},b_{1},b_{2},...,b_{i-1};\\ a_{1},a_{2},a_{3},...,a_{m}
\label{eq2}
\end{aligned}
\end{equation}

The decoder generates output using previously predicted word vectors and source sentence vectors in Eq. \ref{eq1}.

\begin{figure}
\centering
\includegraphics[width=65mm,scale=0.3]{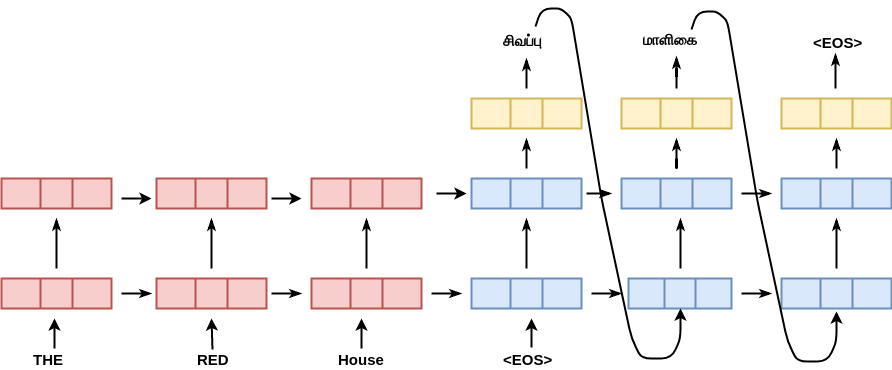}
\caption{Seq2Seq architecture for English-Tamil}
\label{fig:method}
\end{figure}

\subsubsection{Attention Model}
In a basic encoder-decoder architecture, encoder memorizes the whole sentence in terms of vector, and store it in the final activation layer, then the decoder uses that vector to generates the target sentence. This architecture works quite well for small sentences, but for larger sentences, maybe longer than 30 or 40 words, the performance degrades.
To overcome this problem attention mechanisms play an important role. The basic idea behind this is that each time, when the model predicts an output word, it only uses the parts of input where the most relevant information is concentrated instead of the whole sentence. In other words, it only pays attention to some weighted words. Many types of attention mechanisms are used in order to improvise the translation accuracy, but the multi-head self-attention overcomes most of the problems.

\paragraph{Self-attention}
In self-attention architecture \cite{NIPS2017_7181} at every time step of an RNN, a weighted average of all the previous states will be used as an extra input to the function that computes the next state. With the self-attentive mechanism, the network can decide to attend to a state produced many time steps earlier. This means that the latest state does not need to store all the information. The mechanism also makes it easier for the gradient to flow more easily to all previous states, which can help against the vanishing gradient problem.

\begin{figure}[ht]
\centering
\includegraphics[width=60mm,scale=0.3]{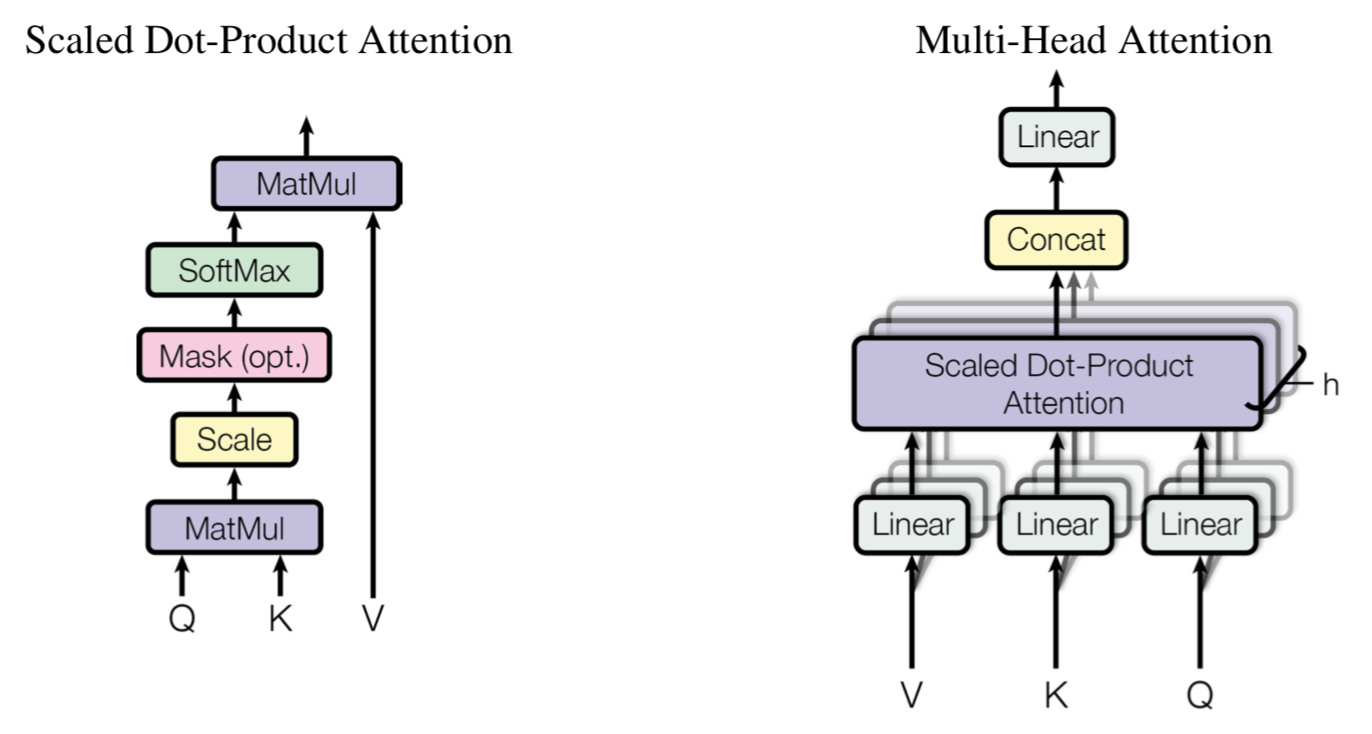}
\caption{Attention model}
\label{attention}
\end{figure}

\paragraph{Multi-Head Attention}
When we have multiple queries q, we can combine them in a matrix Q. If we compute alignment using dot-product attention, the set of equations that are used to calculate context vectors can be reduced as shown in figure 3. Q, K, and V are mapped into lower-dimensional vector spaces using weight matrices and the results are used to compute attention (which we call a Head). In Muti-Head Attention we have h such sets of weight matrices which give us h Heads.

\begin{figure}[ht]
\centering
\includegraphics[width=60mm,scale=0.3]{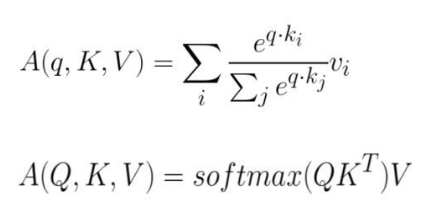}
\caption{Multi-Head Attention}
\label{Multi-Head attention}
\end{figure}

\subsubsection{Word Embedding}
Word embedding is a unique way of representing the word in a vector space such that we can capture the semantic similarity of each word. Each word is represented in hundreds of dimensions. Generally, pre-trained embeddings are used trained on the larger data sets, and with the help of transfer learning, we convert the words from vocabulary to vector. \cite{cho2014properties}.

\subsubsection{Byte Pair Encoding}
BPE \cite{gage1994new} is a data compression technique that replaces the most frequent pair of bytes in a sequence. We use this algorithm for word segmentation, and by merging frequent pairs of charters or character sequences we can get the vocabulary of desired size \cite{sennrich2015neural}. BPE helps in the suffix, prefix separation, and compound splitting which in our case used for creating new and complex words of Malayalam and Tamil language by interpreting them as sub-words units. We used BPE along with pre-trained fast-text word embeddings  \footnote{https://github.com/bheinzerling/bpemb} \cite{heinzerling2018bpemb} for both the languages with the variation in the vocabulary size. In our model, we got the best results with vocabulary size 25000 and dimension 300. 

\paragraph{MultiBPEmb}
MultiBPEmb is a collection of multiple languages subword segmentation models and pre-trained subword embeddings trained on Wikipedia data similar to monolingual BPE. On the contrary, instead of training one segmentation model for each language, here we train a single model and a single embedding for all the languages. We can also create a vocabulary of only two languages, source, and target. It deals with the mixed language sentences (Native language along with English) which are being popular nowadays on social media. Since our sentences were clean it almost produced similar results, with variation in the BLEU score by 0.60 in Tamil and 1.15 in Malayalam. 

\section{Experimentation and Results}
\subsection{Evaluation Metric}
BLEU score is a method to measure the difference between machine translation and human translation \cite{papineni2002bleu}. The approach works by matching n-grams in result translation to n-grams in the reference text, where unigram is a unique token, bigram is a word pair and so on. A perfect match results in a score of 1.0 or 100\%.

\subsection{Dataset}
We obtained the data from different resources such as EnTamV2.0 \cite{biblio}, Opus \cite{TIEDEMANN12.463} and UMC005\cite{JaZeWordOrderIssues2011} .The sentences are of domain news, cinema, bible and movie subtitles. We combined and preprocessed the data of Tamil, Malayalam, Telugu, Bengali, and Urdu. After preprocessing (as described below) and cleaning, the dataset is split into train, test, and validation. Our final dataset is described in table \ref{tab:data}. In our knowledge this is the largest, clean and preprocessed public dataset \footnote{https://github.com/himanshudce/Indian-Language-Dataset} available on the web for general purpose uses. As there is no publicly available dataset to compare various approaches on Indian languages, our datasets can be used to set baseline results to compare with. 

\begin{table}
\centering
\begin{tabular}{|l|l|l|l|l|} 
\hline
ID & ~Language~ & ~ Train  & ~Test~ & ~Dev~   \\ 
\hline
1  & Tamil      & ~183451  & ~2000  & ~1000   \\ 
\hline
2  & Malayalam  & ~548000~ & ~3660~ & ~3000~  \\ 
\hline
3  & Telugu     & ~75000   & ~3897  & ~3000   \\ 
\hline
4  & Bengali    & ~658000  & ~3255  & ~3500   \\ 
\hline
5  & Urdu       & ~36000   & ~2454  & ~2000   \\
\hline
\end{tabular}
\caption{Dataset for Indian Languages}
\label{tab:data}
\end{table}

\subsection{Data Pre-processing}
In the Research works \cite{hans2016improving} \cite{ramesh2018neural} EnTamV2.0 dataset is used. Also, the Opus dataset is a much widely used parallel corpus resource in various researcher's works. However, we observed that in both of these well-known parallel resources there are many repeated sentences, which may results into the wrong results (can be higher or lower) after dividing into train, validation, and test sets, as many of the sentences, occur both in train and test sets. In most of the work, the focus relies on the models without interpreting the data which performs much better on our own test set rather than on general translated sentences. Thus, it is essential to analyses, correct and cleans the data before using it for the experiments. Researchers should also provide a detailed source of the corpus otherwise results can be misleading such as in paper \cite{inproceedings}. We observed the following four important issues in the online available corpus.

\begin{itemize}
\item Sentence repetition with the same source and target.
\item Different translations by the same source.
\item Same translated sentences by different source sentences.
\item Indian language tokenization.
\end{itemize}

To overcome the first issue, we took unique pairs from all the parallel sentences and removed the repeating ones. To tackle the second and third case we removed sentence pairs which were repeated more than twice and the difference between their length are in the window of 5 words. It is because for both of these cases we cannot identify that which source is correct for the same translation and which translated sentence is comes from the same source. We observed that there were some sentences, which were repeating even more than 20 times in the Opus dataset. This confuses the model to learn, identify and capture different features and overfits the model. Though data-augmentation \cite{fadaee-etal-2017-data} can improve the translation results, but in that case, the original data should be pre-processed, otherwise many augmented sentences may appear in both train and test data which leads to higher but wrong BLEU score as it will not work efficiently on new sentences.

For the tokenization of the English language, there are many libraries and frameworks such as (\emph{e.g.}, Perl tokenizer) but these do not work well on the Indian languages, due to the difference between morphological symbols. The word-formation of Indian languages is quite different which we believed can only be handled by either special library for that particular language or by Byte-Pair-Encoding. In the case of BPE, we don't need to tokenize the words which generally leads to better translation results. 

After working on all these minor, but effective pre-processing we got our final dataset. While extracting the data from the web, we also removed sentences with a length greater than 50, known translated words in target sentences, noisy translations, and unwanted punctuations. For the reliability of data, we also took the help of native speakers of these languages.

\subsection{Translator}
We tried various new techniques as described above to get a better intuition of the effects on these two Indian language pairs. Our first model consists of 4 layer Bi-directional LSTM encoder and a decoder with 500 dimensions each along with a vocabulary size of 50,004 words for both source and target. First, we used Bahdanau's attention and Adam optimizer with the dropout (regularization) of 0.3 and the learning rate 0.001. Here we used the 300 dimensional Pre-trained fast text \footnote{https://fasttext.cc/docs/en/crawl-vectors.html} word embeddings for both the languages. Secondly, we used Pre-trained fast text Byte-Pair-Encoding \footnote{https://github.com/bheinzerling/bpemb} with the same attention. In the third model, we changed the attention to multi-head with 8 heads and 6 encoding and decoding layers. It shows an improvement of 1.2 and 6.18 BLEU scores for Tamil and Malayalam respectively. For the final model we used Multilingual fast text pre-trained Byte-pair-Encoddings \footnote{https://nlp.h-its.org/bpemb/multi/} and got our final best results of 9.67 and 25.36 respectively as shown in table \ref{tab:En-ta} and table \ref{tab:En-ml}.

\subsection{Result}
Our results is shown in table \ref{tab:En-ta} and table \ref{tab:En-ml}. For Google translate we used Python API to translate the English sentences and compared the results with our various models. From the test results, It is observed that our model overcomes the OOV (Out of Vocabulary) problem in some cases, and is handy enough to be used in day to day life and official work.

\begin{figure}[ht]
\centering
\textbf{English-Tamil translation models}
\includegraphics[width=80mm,scale=0.5]{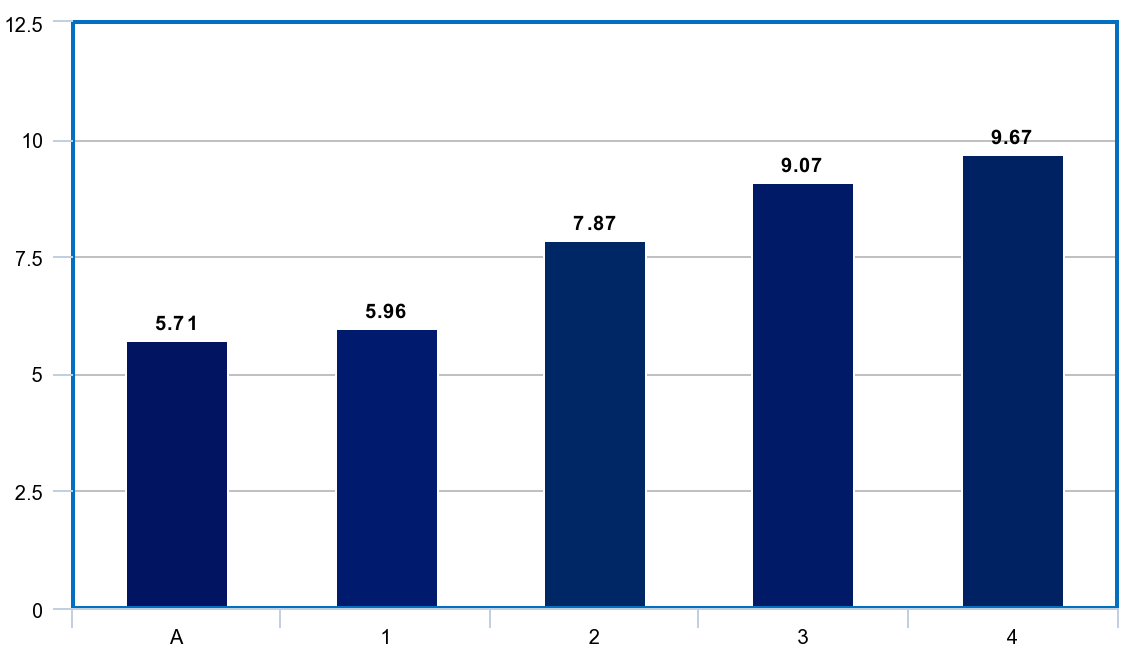}
\caption{English-Tamil model comparison with Google Translator Table\ref{tab:En-ta}}
\end{figure}

\begin{figure}[ht]
\centering
\textbf{English-Malayalam translation models}
\includegraphics[width=80mm,scale=0.5]{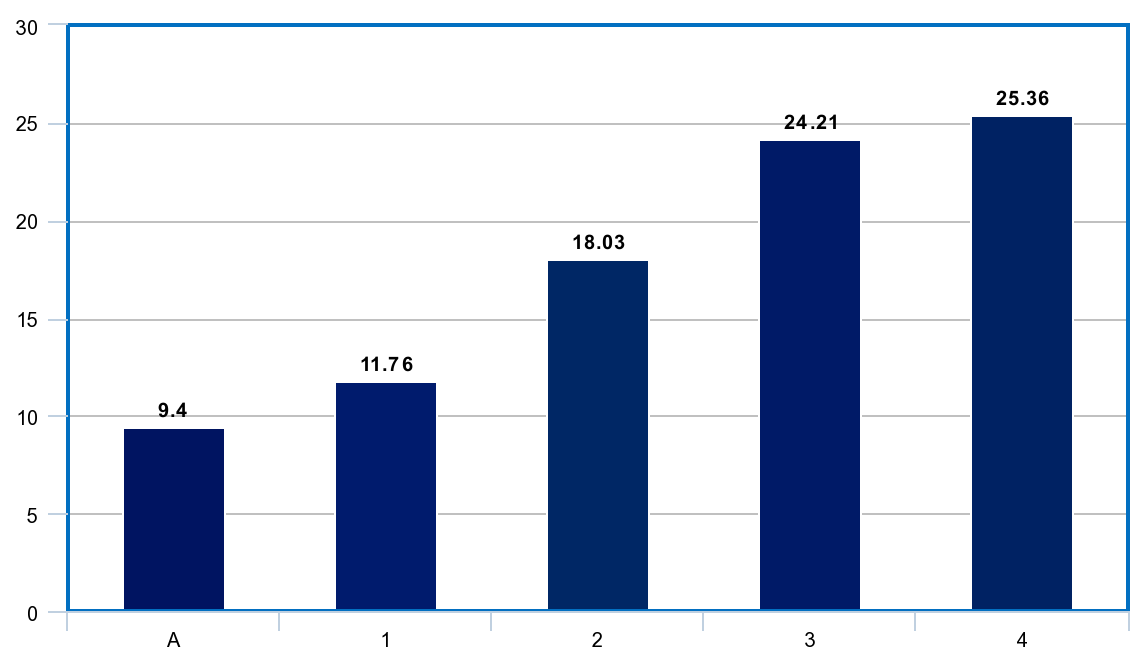}
\caption{English-Malayalam model comparison with Google Translator Table \ref{tab:En-ta}}
\end{figure}

\begin{table}
\centering
\begin{tabular}{|l|l|l|} 
\hline
ID                                            & Model                                                                                                             & \begin{tabular}[c]{@{}l@{}}~BLEU~~\\~Score\end{tabular}  \\ 
\hline
\begin{tabular}[c]{@{}l@{}}\\A\\\end{tabular} & Google Translator                                                                                                 & 5.71                                                     \\ 
\hline
1                                             & \begin{tabular}[c]{@{}l@{}}Bi-LSTM(4-Layers)+ A+\\~Bahdanau Attention + WE\end{tabular}                           & 5.96                                                     \\ 
\hline
2                                             & \begin{tabular}[c]{@{}l@{}}Bi-LSTM(4-Layers)+ A +\\~Bahdanau Attention~+ Pre-BPE(25000)\end{tabular}              & 7.87                                                     \\ 
\hline
3                                             & \begin{tabular}[c]{@{}l@{}}Bi-LSTM(6-Layer)+ A + \\Multi-Head Attention+ Pre-BPE(25000)~\end{tabular}             & 9.07                                                     \\ 
\hline
4                                             & \begin{tabular}[c]{@{}l@{}}Bi-LSTM(6-Layer)+ A + \\Multi-Head self Attention\\+ Pre-MultiBPE(100000)\end{tabular} & 9.67                                                     \\
\hline
\end{tabular}
\caption{English-Tamil model comparison with Google Translator ( A=Adam, WE=Word Embeddings)}
\label{tab:En-ta}
\end{table}

\begin{table}
\centering
\begin{tabular}{|l|l|l|} 
\hline
ID                                            & Model                                                                                                             & \begin{tabular}[c]{@{}l@{}}~BLEU~~\\~Score\end{tabular}  \\ 
\hline
\begin{tabular}[c]{@{}l@{}}\\A\\\end{tabular} & Google Translator                                                                                                 & 9.40                                                     \\ 
\hline
1                                             & \begin{tabular}[c]{@{}l@{}}Bi-LSTM(4-Layers)+ A+\\~Bahdanau Attention + WE\end{tabular}                           & 11.76                                                    \\ 
\hline
2                                             & \begin{tabular}[c]{@{}l@{}}Bi-LSTM(4-Layers)+ A +\\~Bahdanau Attention~+ Pre-BPE(25000)\end{tabular}              & 18.03                                                    \\ 
\hline
3                                             & \begin{tabular}[c]{@{}l@{}}Bi-LSTM(6-Layer)+ A + \\Multi-Head Attention+ Pre-BPE(25000)~\end{tabular}             & 24.21                                                    \\ 
\hline
4                                             & \begin{tabular}[c]{@{}l@{}}Bi-LSTM(6-Layer)+ A + \\Multi-Head self Attention\\+ Pre-MultiBPE(100000)\end{tabular} & 25.36                                                    \\
\hline
\end{tabular}
\caption{BLEU Score of English-Malayalam translated system. \ (A=Adam, B= Bahdanau, WE=Word Embedding)}
\label{tab:En-ml}
\end{table}

\begin{figure}[!h]
\centering
\textbf{Attention Visualization}
\includegraphics[width=80mm,scale=0.5]{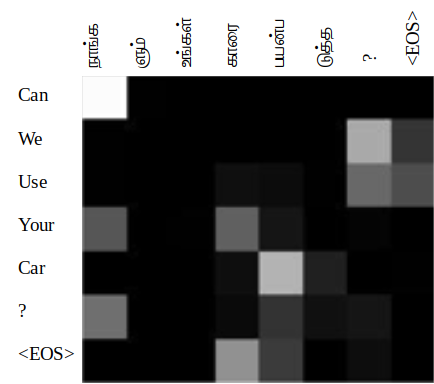}
\caption{Attention visualization of English-Tamil sentence pair from our test data}
\label{fig:attention}
\end{figure}

\subsection{Analysis}
We conducted a survey with ten random sentences from our test data and accumulated the reviews of native Tamil speaking peoples. On comparing the reviews of Google translator and our translator, it was found, that our translation results were better in 60\% cases than the Google translator. The visualization of an Attention can be seen in \ref{fig:attention} of one of the sample sentences from our test data.

\section{Conclusion} \label{conclusion}
In this paper, we applied Neural Machine Translation (NMT) on two of the most difficult Indian language pairs (English-Tamil, English-Malayalam). We addressed the issues of data pre-processing and tokenization. To handle morphology and word complexities of Indian languages we applied pre-trained fast text BPEmb, MultiBPEmb embeddings along with multi-head self-attention which outperformed Google translator with a margin of 3.96 and 15.96 BLEU points respectively. The same approach can be applied to other Indian languages as well. Since the accuracy of our model was fairly good, it can be used for creating English-Malayalam and English-Tamil translation applications that will be very useful in domains like tourism, education and corporate. In the future, we can also explore the possibility to improve the translation results for code-switched languages using MultiBPE and other variations.

\section{Bibliographical References}

\bibliography{references}
\bibliographystyle{lrec}
\end{document}